\documentclass{article}
\usepackage{ijcai22}

\usepackage{times}
\usepackage{soul}
\usepackage{url}
\usepackage[hidelinks]{hyperref}
\usepackage[utf8]{inputenc}
\usepackage[small]{caption}
\usepackage{graphicx}
\usepackage{amsmath}
\usepackage{amsthm}
\usepackage{booktabs}
\usepackage{algorithm}
\usepackage{algorithmic}
\graphicspath{ {./pictures/} }

\title{Composition-aware Graphic Layout GAN for \\Visual-textual Presentation Designs}

\author{
Min Zhou$^1$\footnote{Equal contribution}\and
Chenchen Xu$^{1,2}$\footnotemark[1]\footnote{Work done during an internship at Alibaba Group}\and
Ye Ma$^1$\and
Tiezheng Ge$^1$\and
Yuning Jiang$^1$\And
Weiwei Xu$^2$\footnote{Corresponding author}\\
\affiliations
$^1$Alibaba Group\\
$^2$Zhejiang University\\
\emails
yunqi.zm@alibaba-inc.com,
xuchenchen@zju.edu.cn,
\{maye.my, tiezheng.gtz, mengzhu.jyn\}@alibaba-inc.com,
xww@cad.zju.edu.cn
}

\begin{document}

\maketitle

\begin{abstract}
In this paper, we study the graphic layout generation problem of producing high-quality visual-textual presentation designs for given images. We note that image compositions, which contain not only global semantics but also spatial information, would largely affect layout results. Hence, we propose a deep generative model, dubbed as composition-aware graphic layout GAN (CGL-GAN), to synthesize layouts based on the global and spatial visual contents of input images. To obtain training images from images that already contain manually designed graphic layout data, previous work suggests masking design elements (e.g., texts and embellishments) as model inputs, which inevitably leaves hints of the ground truth. We study the misalignment between the training inputs (with hint masks) and test inputs (without masks), and design a novel domain alignment module (DAM) to narrow this gap. For training, we built a large-scale layout dataset which consists of 60,548 advertising posters with annotated layout information. To evaluate the generated layouts, we propose three novel metrics according to aesthetic intuitions. Through both quantitative and qualitative evaluations, we demonstrate that the proposed model can synthesize high-quality graphic layouts according to image compositions. The data and code will be available at https://github.com/minzhouGithub/CGL-GAN.
\end{abstract}

\section{Introduction}
\begin{figure}
    \centering
    \includegraphics[width=8.5cm]{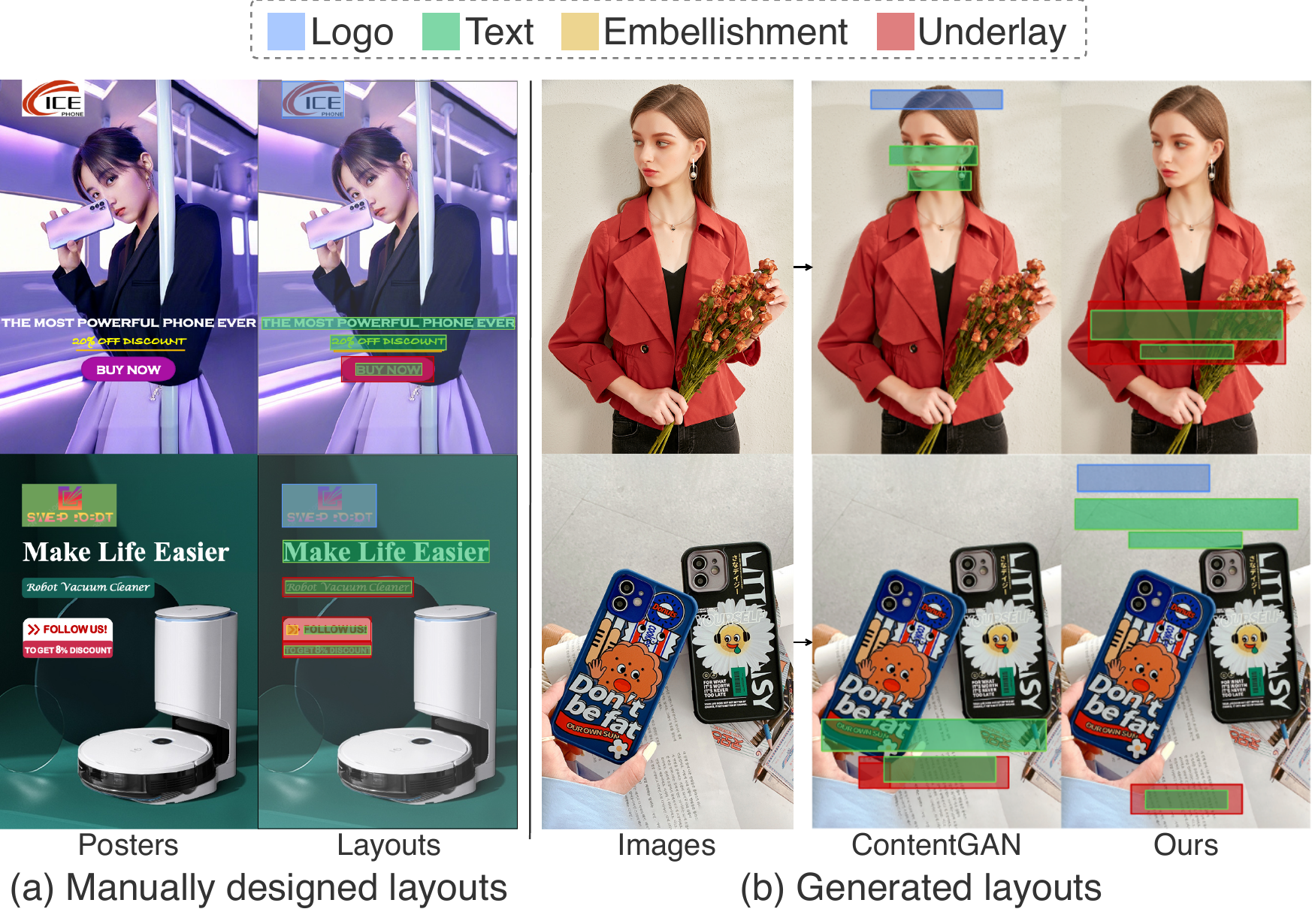}
    \caption{(a) Poster layouts annotated from manually designed posters; (b) poster layouts generated by automatic algorithms of baseline~\protect\cite{contentGAN} and ours.}
    \label{fig:teaser}
\end{figure}

Images with overlaid texts and embellishments, also known as visual-textual presentations~\cite{yang2016automatic}, are becoming more ubiquitous in the form of advertising posters (see Fig.~\ref{fig:teaser}(a)), magazine covers, etc. 
In this paper, we generally call them posters. To automatically generate visually pleasing posters with input images, determining what and where to put graphic elements on images (namely graphic layout design) plays an important role. 

Layout design of posters is challenging in the sense that it should consider not only graphic relationships but also image compositions, and has attracted continuous research efforts. 
Recently, deep-learning-based algorithms~\cite{DBLP:layoutGAN,attrGAN,DBLP:conf/cvpr/ArroyoPT21} have been proposed to automatically generate layouts. However, these methods focus more on graphic relationships while ignoring image contents, which is essential for visual-textual posters.

ContentGAN~\cite{contentGAN} is the first to introduce image semantics in layout generation. Benefiting from content information, it can yield high-quality layouts for magazine pages. However, we note the neglect of image compositions, especially the spatial information, in this work. Such problems can negatively impact the subject presentations, text readability, or even the visual balance of the whole poster. As shown in Fig.~\ref{fig:teaser}(b), human faces or products are occluded, and texts are wrongly assigned leaving a large blank region. To tackle these problems and better learn the relationship between images and overlaid elements, we combine a multi-scale CNN and a transformer to take full advantage of image compositions. A structurally similar discriminator is involved to distinguish whether the images and layouts are matched.

To overcome the lack of data in poster layout design, we construct a large-scale dataset. Considering the high cost of acquiring image-layout pairs by manual design, we just collect posters and images from websites following~\cite{contentGAN}. The posters are labeled for training as shown in Fig.~\ref{fig:teaser}(a) and the images are for test (the first row of Fig.~\ref{fig:teaser}(b)). 

Note that there is an obvious domain gap between training posters and test images due to graphic elements. To address this, ContentGAN masks graphic elements on training data\footnote{\url{https://xtqiao.com/projects/content_aware_layout}} and uses a global pooling after the frozen image-feature extractor, by which image compositions are lost for the change of domain adaptation. Here we design a domain alignment module (DAM), which consists of an inpainting subnet and a saliency detection subnet, to narrow the domain gap and maintain image compositions well.

Existing metrics~\cite{DBLP:layoutGAN,DBLP:conf/iccv/JyothiDHSM19,DBLP:conf/cvpr/ArroyoPT21} only consider the relationship between graphic elements and ignore the one between graphic elements and image compositions. So, we additionally use user studies and three novel composition-relevant metrics to verify our method. The experiments demonstrate that our method can yield high-quality layouts for images and outperform other state-of-the-art methods. Besides, our method can produce layouts to fit input user constraints (incomplete layouts). 

In conclusion, our main contributions are as follows:
 \begin{itemize}
     \item We propose a novel method to generate composition-aware graphic layouts for visual-textural posters. First, a domain alignment module (DAM) is designed to run the training process without obtaining complete image-poster-annotation pairs (poster-annotation pairs are enough). Secondly, the composition-aware layout generator is proposed to model the relationship between image compositions and graphic layouts. 
     \item We contribute a large layout dataset including all different kinds of promotional products and delicate designs. To the best of our knowledge, it is the first large-scale dataset in advertising poster layout design.
     \item On this dataset, we demonstrate that our method effectively tackles the image-composition-aware layout generation problem without paired images and layouts. And the experiments show our model outperforms state-of-the-art methods. In addition, our model is capable of generating layouts with input constraints. 
 \end{itemize}

\begin{figure*}
    \centering
    \hspace{-1cm}\includegraphics[width=18.1cm]{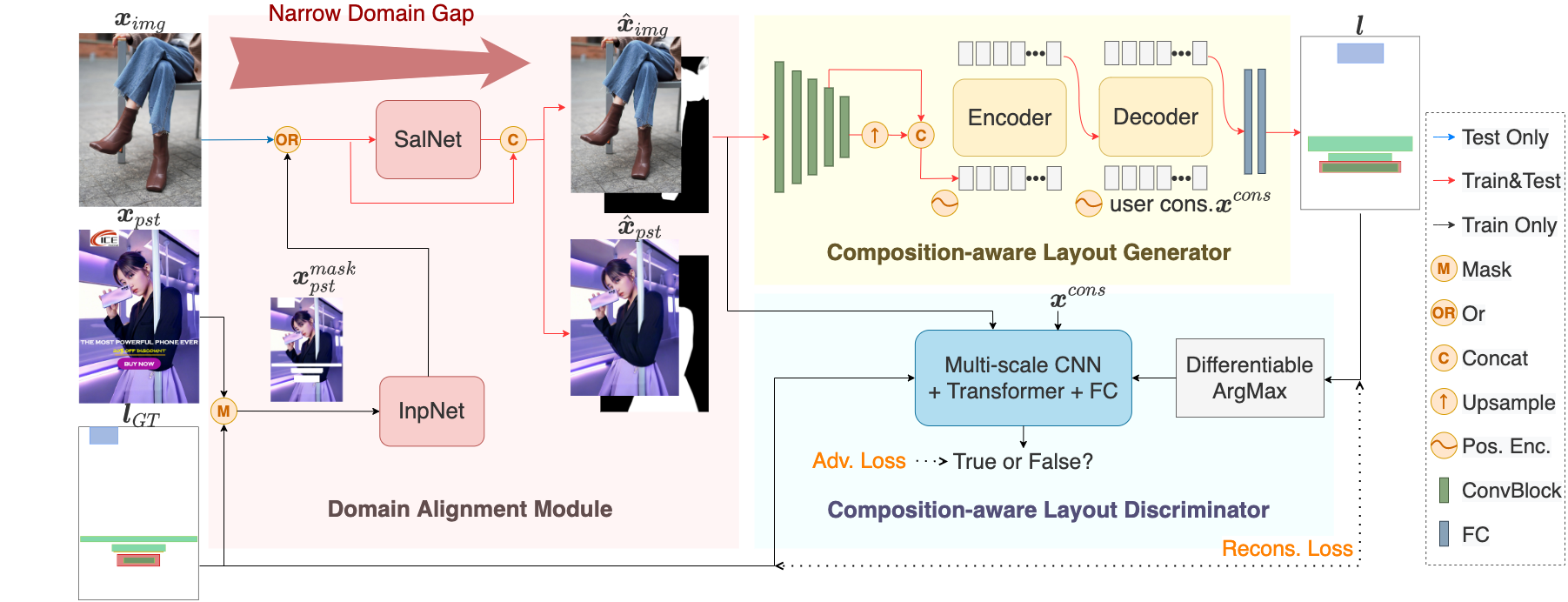}
    \caption{{\bf The framework of our model.} The training and test data first go through the DAM to be less distinguishable in domain. Sequentially, a generator, consisting of a multi-scale CNN and a transformer, is applied to yield layouts based on outputs of DAM and user constraint layouts. Besides, a discriminator structurally similar to the generator is used for training. The whole model is trained with a reconstruction loss and an adversarial loss.}
    \label{fig:2}
\end{figure*}

\section{Related Work}
In automatic layout generation, early works mainly rely on templates~\cite{DBLP:journals/tog/JacobsLSBS03} or heuristic methods~\cite{DBLP:conf/chi/KumarTAK11,DBLP:journals/tog/CaoCL12,DBLP:journals/tvcg/ODonovanAH14}. They require professional knowledge and often fail to yield flexible and various layouts limited to their hand-crafted rules.

As deep learning develops, LayoutGAN~\cite{DBLP:layoutGAN}, LayoutVAE~\cite{DBLP:conf/iccv/JyothiDHSM19} and VTN~\cite{DBLP:conf/cvpr/ArroyoPT21} appear to produce layouts from noise. Meanwhile, some conditional layout generation methods have been proposed~\cite{attrGAN,DBLP:conf/eccv/LeeJELG0Y20,DBLP:conf/cvpr/YangFYW21,DBLP:conf/mm/KikuchiSOY21}. But all these methods concentrate on learning the internal relationship of graphic elements and ignore visual content.

ContentGAN~\cite{constrainGAN} uses semantic visual information to generate layouts for magazine pages. But as mentioned earlier, due to the lack of spatial information and detailed features, it encounters problems when applied to poster layout generation, such as the occlusion of subjects. In contrast, by mining image-composition information, our method can better understand image contents and output high-quality layouts for posters.
 
\section{Dataset and Representation}
We collect all data from e-commerce platforms (advertising posters for training and images from product pages for test). 

Posters in this dataset are diverse, not only in terms of the product category (e.g., cosmetics, electronics, clothing, etc.), but also the display format of products (e.g., size, quantity, background, etc.).

As Fig.~\ref{fig:teaser} shows, we classify graphic elements into four fine categories: logos, texts, underlays and embellishments. All elements are manually labeled with their categories and bounding boxes. After the manual double check, we finally get 60,548 poster-layout pairs to train. As for test, we select 1,000 pure images from all types of products.

We represent a graphic layout as a variable-length set as ${\{\boldsymbol{e}_1, \boldsymbol{e}_2, ..., \boldsymbol{e}_N\}}$ (${N}$ is the number of elements). Each element $\boldsymbol{e}_i$ is composed of a class ${\boldsymbol{c}_i}$ and a bounding box ${\boldsymbol{b}_i}$. ${\boldsymbol{c}_i}$ is in one-hot form and ${\boldsymbol{b}_i}$ contains the central position and size of the box. For numerical optimization, each dimension of ${\boldsymbol{b}_i}$ is scaled between ${[0, 1]}$ divided by image width or height.

\section{Composition-aware Graphic Layout GAN}
In this section, we will introduce the main parts of our model sequentially along the data flow during training.

\subsection{Domain Alignment Module}

As mentioned earlier, the inputs are posters for training but changed to images for test. It is obvious in Fig.~\ref{fig:teaser} that graphical elements in posters inevitably leave hints for the ground truth, and also lead to a gap between training/test data.

Aimed at narrowing the domain gap, a Domain Alignment Module (DAM) is proposed and illustrated in Fig.~\ref{fig:2}. For a poster ${\boldsymbol{x}_{pst}}$ in training, we first get a masked poster ${\boldsymbol{x}^{mask}_{pst}}$ guided by its layout ${\boldsymbol{l}_{GT}}$ and utilize a pretrained inpainting Net~\cite{DBLP:conf/wacv/SuvorovLMRASKGP22} (InpNet) to obtain repaired result ${\boldsymbol{x}^{inp}_{pst}}$. Besides, we use a pretrained salience detection net (SalNet)~\cite{DBLP:journals/corr/abs-1911-05942} to point out subject locations and help subsequent networks better understand image compositions. So ${\boldsymbol{x}^{inp}_{pst}}$ is sent into SalNet to output ${\boldsymbol{x}^{sal}_{pst}}$ and these two are concatenated as ${\hat{\boldsymbol{x}}_{pst}}$. While in test, an image ${\boldsymbol{x}_{img}}$ is just sent into SalNet to get ${\boldsymbol{x}^{sal}_{img}}$ and concatenated with it to form ${\hat{\boldsymbol{x}}_{img}}$. Through inpainting and salience detection, the marks of graphic elements are almost eliminated and the domain gap between ${\hat{\boldsymbol{x}}_{pst}}$ and ${\hat{\boldsymbol{x}}_{img}}$ is much smaller than that between ${\boldsymbol{x}_{pst}}$ and ${\boldsymbol{x}_{img}}$.

To reduce the differences caused by inpainting artifacts and salience detection, we apply gaussian blur to ${\boldsymbol{x}^{inp}}$ and morphological processes to ${\boldsymbol{x}^{sal}}$ before the concatenation. 

\subsection{Composition-aware Layout Generator}
In order to model the relationship between image compositions and layout elements, the layout generator consists of three parts: a multi-scale CNN backbone extracting image features, a transformer implicitly learning layout generating rules, and two fully connected layers (FCs) respectively predicting layout element classes and box coordinates.

As the output from DAM, ${\hat{\boldsymbol{x}}}$ is fed into the CNN backbone, a ResNet50~\cite{DBLP:conf/cvpr/HeZRS16} whose input channels are changed to four, and the part after its final convolutional layer is removed. For the reason that image compositions not only mean high-level semantics such as subject locations but also include lower-level features like region complexity, we introduce a multi-scale strategy on the last two convolutional blocks following FPN~\cite{fpn}. There is a slight difference that we do not generate layouts on each scale separately like detection networks often do, but concatenate the fused and upsampled features as one. We denote $\boldsymbol{F}_j$ the feature maps of the ${j}$-th convolutional block. Then multi-scale features can be computed as
\begin{equation}
\begin{aligned}
\boldsymbol{F}^{'}_j = \mathbf{Conv_{11}}(\boldsymbol{F}_j)\;;\quad\boldsymbol{F}^{up}_j = \mathbf{Upsample}(\boldsymbol{F}^{'}_j)\;;
\\
\boldsymbol{F}^{fused}_j = \mathbf{Cancat}(\boldsymbol{F}^{up}_j, \mathbf{Conv_{33}}(\boldsymbol{F}^{up}_j+\boldsymbol{F}^{'}_{j-1}))
\end{aligned}
\end{equation}
where $\mathbf{Conv_{11}}$, $\mathbf{Conv_{33}}$, $\mathbf{Upsample}$ and  $\mathbf{Cancat}$ are network operation of  convolution,  up-sampling and concatenation respectively.

Though higher resolution features can provide more details, we empirically find fusing features from the last two blocks yields good results and is more efficient to train.

The multi-scale features are projected into ${d}$ channels and flattened by channel as the beginning of the transformer encoder. The encoder uses a standard transformer architecture and further refines the image features. The decoder takes in an initial layout ${\boldsymbol{x}^{cons}}$ (a layout constraint, if any else an empty one) and utilizes cross-attentions to learn the relationship between image compositions and graphic layouts. Also, the internal relationship of graphic elements is built through self-attentions of the decoder.
The encoder and decoder both have six layers and the hidden dims are ${d}$. Position encodings are added both in the encoder and decoder.

In the end, the decoder features of each element are taken into two FCs to predict the corresponding class ${\boldsymbol{c}}$ and box coordinates ${\boldsymbol{b}}$ logits. The weights of FCs for all elements are shared. The final class and box results are calculated using softmax or sigmoid functions.

For convenience, the generator is non-autoregressive. Considering the number of elements ${N}$ is uncertain, our model also introduces non-objects and bipartite matching to calculate reconstruction loss $L_{rec}$ following~\cite{DBLP:conf/eccv/CarionMSUKZ20}. That is to say, all layouts are padded to the max length and padding elements are non-objects with all-zero coordinates.

\begin{table*}[ht]
    \centering
    \setlength{\tabcolsep}{2mm}{
    \centering
    \begin{tabular}{lrrrr|rrrr|rrr}
    \toprule
         Model    &$P^{*}_{qs}\uparrow$ &$P^{*}_{best}\uparrow$ &$P_{qs}\uparrow$ &$P_{best}\uparrow$ &$R_{occ}\uparrow$ &$R_{com}\downarrow$ &$R_{sub}\downarrow$ &$R_{shm}\downarrow$ &$R_{ove}\downarrow$ &$R_{und}\uparrow$ &$R_{ali}\downarrow$\\ 
    \midrule
         ContentGAN     &62.00 &32.37 &43.16 &25.25 &93.4 &45.59 &1.143 &17.08 &0.0397 &0.8626 &0.0071 \\
         Ours     &75.67 &67.63 &70.50 &57.88 &99.7 &34.01 &0.816 &14.77 &0.0256 &0.9413 &0.0098 \\
    \bottomrule
    \end{tabular}}
    \caption{{\bf Comparison with ContentGAN.} The user study, composition-relevant, and graphic metrics are listed from left to right. $R_{occ}$ means the ratio of non-empty layouts predicted by models.}
    \label{tab:contentGAN}
\end{table*}

\begin{table*}[ht]
    \centering
    \setlength{\tabcolsep}{1.8mm}{
    \centering
    \begin{tabular}{lrrrr|rrrr|rrr}
    \toprule
         Model   &$P^{*}_{qs}\uparrow$ &$P^{*}_{best}\uparrow$ &$P_{qs}\uparrow$ &$P_{best}\uparrow$ &$R_{occ}\uparrow$ &$R_{com}\downarrow$ &$R_{sub}\downarrow$ &$R_{shm}\downarrow$ &$R_{ove}\downarrow$ &$R_{und}\uparrow$ &$R_{ali}\downarrow$\\ 
    \midrule
         LayoutTransformer      &54.33 &22.59 &37.67 &22.32 &100.0 &40.92 &1.310 &21.08 &0.0156 &0.9516 &0.0049 \\
         LayoutVTN    &53.33 &24.44 &31.42 &16.86 &99.9 &41.77 &1.323 &22.21 &0.0130 &0.9698 &0.0047 \\
         Ours     &76.00 &52.96 &75.08 &60.82 &99.7 &34.01 &0.816 &14.77 &0.0256 &0.9413 &0.0098 \\
    \bottomrule
    \end{tabular}}
    \caption{{Comparison with content-agnostic methods.}}
    \label{tab:vtn}
\end{table*}

\subsection{Composition-aware Layout Discriminator}
The discriminator has a similar structure to the above generator but two points need to be noted.

The initial layout ${\boldsymbol{x}^{cons}}$ is no longer alone but concatenated with a ground truth or predicted layout. Further, to eliminate the numerical differences between ground truth and predicted layouts, a differentiable argmax is applied on predicted layouts, in which gradients are just copied to the generator outputs. And the coordinates of predicted non-objects are reset to zero.

Because the task for discriminator is much easier, its CNN backbone is from Resnet18 and the number of layers in the encoder and decoder are both reduced to four. Only one FC is set behind the decoder judging whether the input image-layout pairs are true or not. We choose hinge loss~\cite{DBLP:journals/corr/LimY17} $L_{adv}$ to train the discriminator. And the overall loss is the weighted sum of $L_{adv}$ and $L_{rec}$.

In addition, we do not choose a wire-frame discriminator like ~\cite{DBLP:layoutGAN} because we empirically found that this relation discriminator with transformers performs better, the same as~\cite{DBLP:conf/mm/KikuchiSOY21}.

\begin{figure*}[ht]
    \centering
    \hspace{-0.2cm}\includegraphics[width=17.7cm]{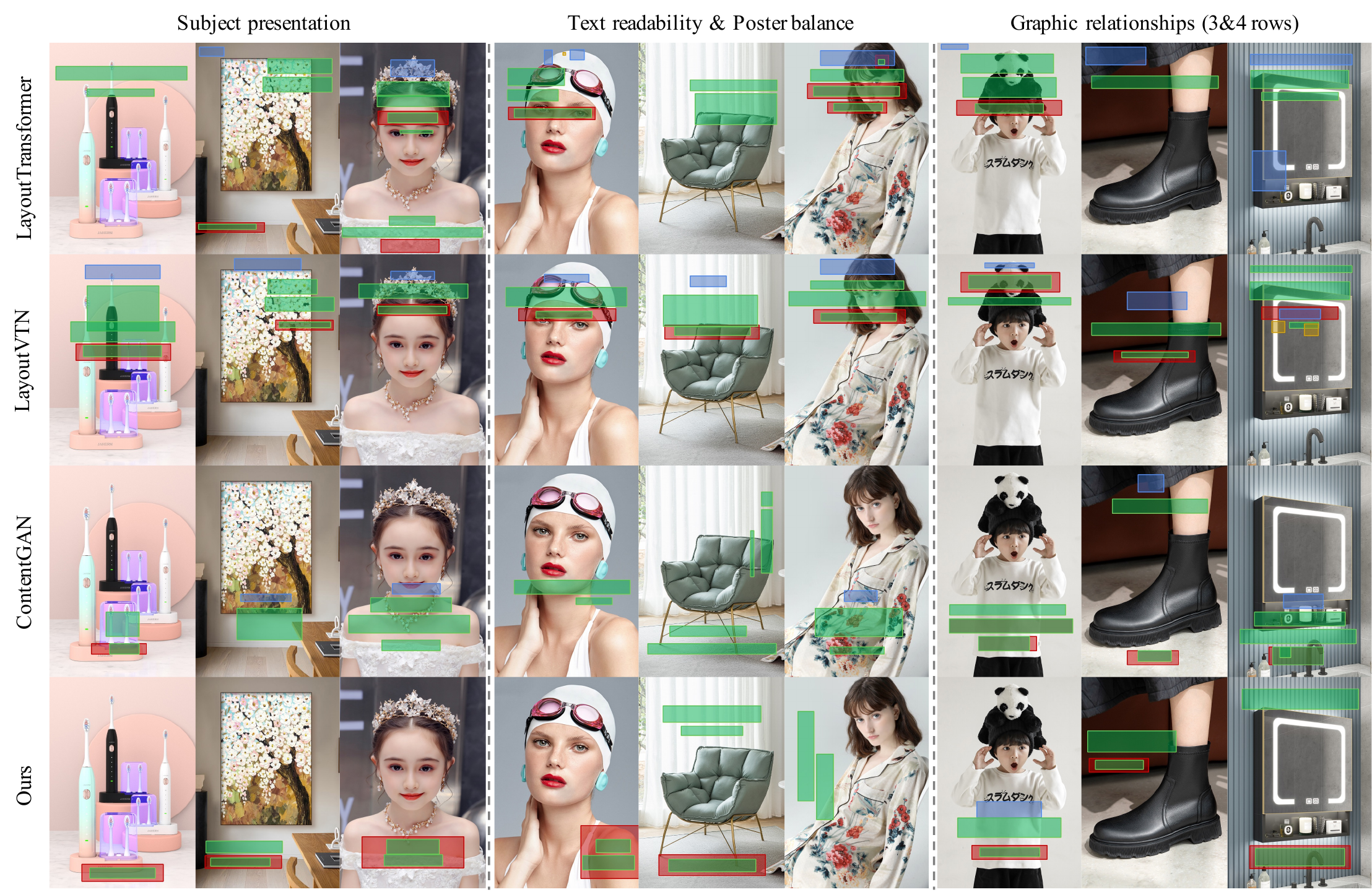}
    \caption{{\bf Qualitative evaluation for different models.} Layouts in a column are conditioned with the same image. And those in a row are from the same method. This figure intuitively shows that our results are better, especially in terms of what top texts hint.}
    \label{fig:3}
\end{figure*}

\section{Metrics}
To better evaluate poster layout generation conditioned by images, we use three kinds of quantitative measures.

\paragraph{User study.}To comprehensively assess layout eligibility, we include evaluation items such as the product presentation, the visual balance of posters, and the overlap of elements. For each test, we prepare shuffled results from different methods on 60 random test images. We involve two groups (five professional and twenty novice designers). Each designer is asked to judge whether each layout is eligible, and pick the best one among compared layouts for the same images. We compute the percentage of layouts reaching the quality standard $P_{qs}$ and of being chosen as the best ones $P_{best}$ ($P^{*}_{qs}$ and $P^{*}_{best}$ for the professional group) for each method.

\paragraph{Composition-relevant measures.}Since there are no existing composition- or content-relevant measures for poster layouts, we propose three novel ones according to design rules.

 \begin{itemize}
    \item Readability and visual balance. When making posters by hand, designers prefer to place texts without underlays on relatively flat regions to ensure text readability and visual balance. Inspired by this, we propose $R_{com}$ to measure the layout quality. Firstly, we convert a test image to its grayscale, and calculate x- and y-direction gradients using the Sobel operator. The final gradient of a pixel is the root mean square of gradients for both directions. We define $R_{com}$ as the average gradients of the pixels covered by predicted text-only elements. Intuitively, a lower $R_{com}$ means better performance.

    \item Presentation of subjects. Attractive advertising posters should highlight the promoted products. For this, we figured out two evaluation ways, $R_{sub}$ and $R_{shm}$. To calculate $R_{sub}$, we get attention maps of promoted products (queried by their category tags extracted on product pages) on test images by CLIP\footnote{\url{https://github.com/hila-chefer/Transformer-MM-Explainability}}~\cite{DBLP:conf/icml/RadfordKHRGASAM21,DBLP:journals/corr/abs-2103-15679} and sum the attention values within layout regions. To calculate $R_{shm}$, we respectively feed the salient images with or without layout regions masked into a pretrained VGG16~\cite{DBLP:journals/corr/SimonyanZ14a}, and calculate ${L_2}$ distance between their output logits. $R_{sub}$ and $R_{shm}$ can reflect the occlusion levels of key subjects and the lower ones are better.
 \end{itemize}
 
\paragraph{Graphic measures.}
Composition-irrelevant graphic measures, such as overlap $R_{ove}$ and alignment $R_{ali}$ of layout elements, are also introduced as in~\cite{attrGAN,DBLP:conf/cvpr/ArroyoPT21}. But here we exclude underlays and embellishments considering the poster nature shown in Fig.~\ref{fig:teaser}. Additionally, because underlays serve for other elements in posters and do not appear alone, we define a metric called underlay overlap $R_{und}$. For each underlay, we find all elements in other categories and intersected with it. Then we compute the overlap ratio of every element and choose the max one as $R_{und}$. It is positively associated with layout qualities. When $R_{und}$ reaches one, the underlay includes at least one other element.

\begin{figure}[ht]
    \centering
    \includegraphics[width=8cm]{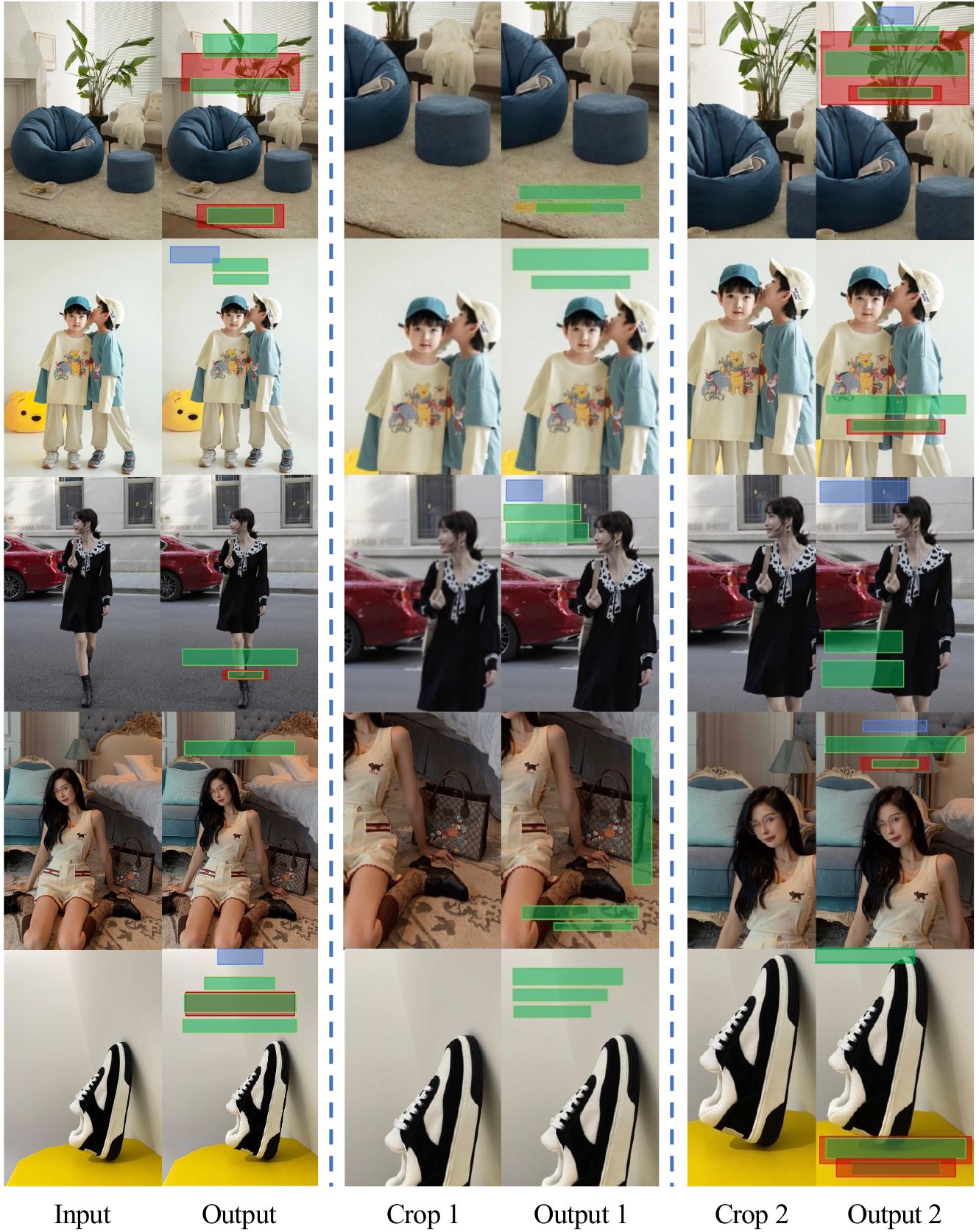}
    \caption{Layouts vary along with the input images (Crop1 and Crop2 are different parts of Input).}
    \label{fig:4}
\end{figure}

\begin{figure}[ht]
    \centering
    \includegraphics[width=8.1cm]{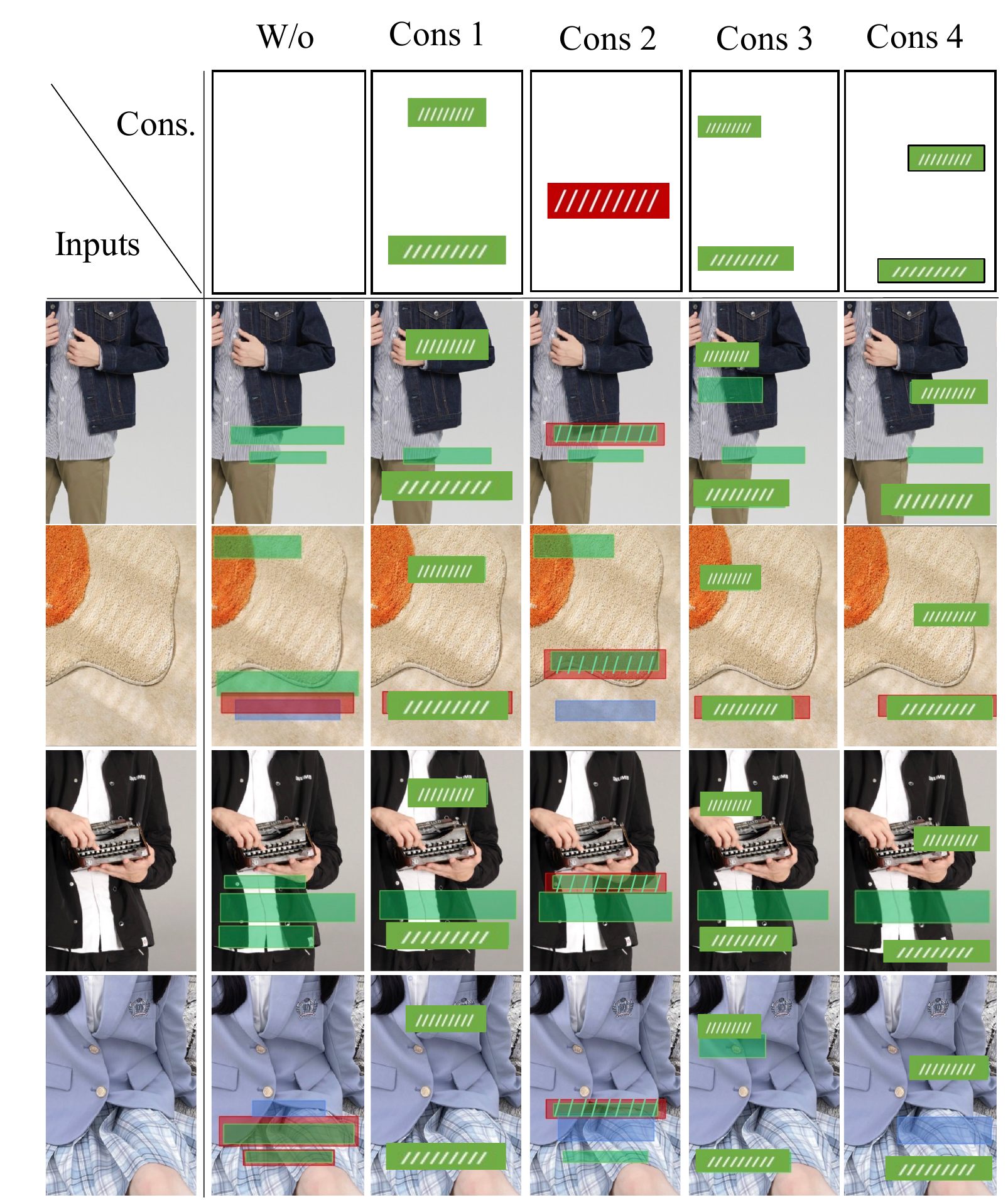}
    \caption{Layouts under user constraints. Constraints are marked by shadowed rectangles.}
    \label{fig:5}
\end{figure}

\section{Experiments}
In this section, we discuss the quantitative and qualitative performance of models. 
\subsection{Implementation Details}

We implement our CGL-GAN in PyTorch and use Adam optimizer~\cite{DBLP:journals/corr/KingmaB14}. The initial learning rates for the generator and discriminator are ${10^{-4}}$ and ${10^{-3}}$ respectively (${10^{-5}}$ and ${10^{-4}}$ for their corresponding backbones). The whole model is trained for 300 epochs with batch size 128 and all learning rates are reduced by a factor of 10 in the last 100 epochs. The weight of $L_{adv}$ linearly increases from zero to one after a 50-epoch warm up. The weight of $L_{rec}$ is kept as one during the whole training. For efficiency, the posters and images are resized as $240\times350$ for models. The hidden dim in transformers is set to 256. 

\subsection{Comparison with SOTA Methods}
\paragraph{Content-aware methods.} As mentioned above, ContentGAN generates layouts considering image contents and is our main comparison here. Based on the released codes \footnote{\url{https://xtqiao.com/projects/content_aware_layout}}, we add content-feature extraction and post-process parts (to be fair, except align and sampling strategy) to reimplement ContentGAN. The quantitative results can be seen in Tab.~\ref{tab:contentGAN}. Ours is much better in the user study and composition-relevant metrics, which indicates that CGL-GAN improves the relationship modeling between image compositions and layouts. 

The left part of Fig.~\ref{fig:3} shows CGL-GAN captures the location of the displayed subjects, so its predicted layout helps rather than effects the display of the products or models. 
The middle part shows CGL-GAN has learned some practical aesthetic rules that texts are better put on flat regions for readability and visual balance of posters, and underlays should be collocated with text when the backgrounds are complex. And the third part implicates that our model also outperforms in handling the internal relationship between graphic elements. More qualitative evaluations are shown in the supplementary.

\paragraph{Content-agnostic methods.} Moreover, we also do our best to implement recent content-agnostic methods~\cite{DBLP:conf/cvpr/YangFYW21,DBLP:conf/cvpr/ArroyoPT21} and use nucleus sampling to generate various results for comparison. As shown in Tab.~\ref{tab:vtn} and Fig.~\ref{fig:3}, it is not surprising to see our model has superiority in user study and composition-relevant metrics. But ours works worse on graphic metrics, especially the alignment. We infer the reason is that the task of these content-agnostic methods is simpler, and they are not influenced by changing images.

\begin{table}[!t]
    \centering
    \setlength{\tabcolsep}{0.55mm}{
    \scalebox{0.945}{
    \begin{tabular}{lcccc|ccc}
    \toprule
         Input    &$R_{occ}\uparrow$ &$R_{com}\downarrow$ &$R_{sub}\downarrow$ &$R_{shm}\downarrow$ &$R_{ove}\downarrow$ &$R_{und}\uparrow$ &$R_{ali}\downarrow$\\
    \midrule
         $\boldsymbol{x}^{mask}$      &2.5 &- &- &- &- &- &-\\
         $\boldsymbol{x}^{sal}$      &\underline{99.6} &39.99 &0.976 &17.82 &0.029 &\underline{0.923} &\bf0.0067\\
         $\boldsymbol{x}^{inp}$     &99.5 &\underline{34.72} &\underline{0.826} &\bf{13.58} &\bf0.021 &0.914 &0.0134\\
         $\hat{\boldsymbol{x}}$      &\bf99.7 &\bf34.01 &\bf{0.816} &\underline{14.77} &\underline{0.026} &\bf0.941 &\underline{0.0098}\\
    \bottomrule
    \end{tabular}
    }
    }
    \caption{{Quantitative ablation study on DAM.}}
    \label{tab:DAM}
\end{table}

\begin{table}[!t]
    \centering
    \setlength{\tabcolsep}{0.6mm}{
    \scalebox{0.945}{
    \begin{tabular}{lcccc|ccc}
    \toprule
         Struc.    &$R_{occ}\uparrow$ &$R_{com}\downarrow$ &$R_{sub}\downarrow$ &$R_{shm}\downarrow$ &$R_{ove}\downarrow$ &$R_{und}\uparrow$ &$R_{ali}\downarrow$\\
    \midrule
         CGD     &97.7  &41.36 &1.180     &19.16 &\bf0.025  &0.821   &\bf0.0075\\
         Ours*    &\bf{99.7}  &\underline{35.13} &\underline{0.897}   &\underline{15.99} &0.044  &\bf{0.961}   &\underline{0.0084}\\
         Ours     &\bf99.7 &\bf34.01 &\bf{0.816} &\bf14.77 &\underline{0.026} &\underline{0.941} &0.0098\\
    \bottomrule
    \end{tabular}
    }
    }
    \caption{{\bf Quantitative ablation study on the generator design.} CGD represents ContentGAN with DAM and Ours* means our method without a multi-scale strategy.}
    \label{tab:composition}
\end{table}

\subsection{Ablations}
\paragraph{Effects of DAM.} In addition to narrowing the domain gap between training and test, DAM also prepares powerful input for the subsequent generator. To verify its effectiveness, we use various inputs to retrain the generator and the results are shown in Tab.~\ref{tab:DAM}. We have tried to input the white-patched poster (${\boldsymbol{x}^{mask}}$) follow~\cite{contentGAN}. In this way, the trained generator degenerates into a white-patch detector, which hardly produces layouts during test. It means that the composition-aware generator cannot handle the data gap without DAM. We also separately use the salient map (${\boldsymbol{x}^{sal}}$) and the inpainted image (${{\boldsymbol{x}}^{inp}}$) as input. As the second row shows, salient-map-only inputs bring a large decrease in composition-relevant metrics for lacking color and textual information. The comparison between the third and fourth rows shows that although only using inpainted images can achieve comparable performance in composition-relevant metrics, the input prepared by DAM (${\hat{\boldsymbol{x}}}$) can make the generator more stable in graphic metrics. We attribute this improvement to the introduction of salient maps.

\paragraph{Effects of generator design.} Aiming at verifying whether the aforementioned generator design helps our model understand image compositions and learn relationships between images and layouts, two comparisons are conducted: ContentGAN equipped with DAM and our generator without a multi-scale strategy. The way for ContentGAN to use DAM is similar to handling multi-image inputs in~\cite{contentGAN}. As can be seen in Tab.~\ref{fig:4}, ContentGAN with DAM still performs poorly, which inversely means that the generator design is indispensable to our model to understand images and generate qualified layouts. The multi-scale strategy helps improve composition-relevant metrics, and provides pros and cons for graphic metrics. 

What is more, as shown in Fig.~\ref{fig:4}, we transform the images and observe that output layouts will change accordingly, rather than just roughly referring to subject categories. And no matter how the subjects are scaled and moved, our model can keep avoiding occlusion on the key regions of subjects (e.g. the human face on the 2nd and 3rd rows) and add underlays on complex regions.

\subsection{Layout Generation Under User Constraints}
When trained with randomly selected partial layouts, from Fig.~\ref{fig:5}, we can see that our model can output various and reasonable layouts according to user constraints on images. Interestingly, in the 4th column, the model knows to put other elements on underlays. And in reverse, the 3rd row shows that it can also add underlays for other elements.

\section{Conclusion}
In this paper, we focus on layout design for visual-textual presentations and propose a novel generative framework named CGL-GAN to handle two main problems: relationship modeling between image compositions and layouts, and domain gap elimination. We contribute a large ad layout dataset and three novel metrics to verify the effectiveness of our model. Future work will focus on predicting more attributes of graphic elements, such as the color of texts.

\section*{Acknowledgements}

This work is partially supported by Alibaba Group through Alibaba Innovation Research Program.

\bibliographystyle{named}
\bibliography{ijcai22}

\end{document}